\documentclass[11pt]{article}
\usepackage[utf8]{inputenc}
\usepackage[T1]{fontenc}
\usepackage{lmodern}
\usepackage[margin=1in]{geometry}
\usepackage{amsmath}
\usepackage{amssymb}
\usepackage{graphicx}
\usepackage{booktabs}
\usepackage[colorlinks=true,linkcolor=blue,citecolor=blue,urlcolor=blue]{hyperref}
\usepackage{microtype}
\usepackage{natbib}
\usepackage{float}

\title{When Correct Isn't Usable: Improving Structured Output\\Reliability in Small Language Models}
\author{%
Cosimo Galeone\\
\texttt{cosimo.galeone@alomana.com}\\
Alomana, Grottaglie, Italy
\and
Minsu Park\\
\texttt{minsu.park@alomana.com}\\
Alomana, Grottaglie, Italy
\and
Giuseppe Ettorre\\
\texttt{ge@alomana.com}\\
Alomana, Grottaglie, Italy
\and
Daniele Ligorio\\
\texttt{dl@alomana.com}\\
Alomana, Grottaglie, Italy
}
\date{}

\begin{document}
\maketitle

\begin{abstract}
Deployed language models must produce outputs that are both correct and format-compliant. We study this structured-output reliability gap using two mathematical benchmarks---GSM8K and MATH---as a controlled testbed: ground truth is unambiguous and the output contract is strict (JSON with required fields). We evaluate three 7--9B models under five prompting strategies and report \emph{output accuracy}---the joint event of mathematical correctness and valid JSON structure---as the primary metric. A systematic format failure emerges: NAIVE prompting (no system prompt) achieves up to 85\% task accuracy on GSM8K but 0\% output accuracy across all models and datasets. REFERENCE prompting (a minimal hand-written JSON format prompt) fares little better, yielding 0\% output accuracy for two of four models tested. Constrained decoding enforces syntactic validity but incurs 3.6$\times$--8.2$\times$ latency overhead and in several settings degrades task performance substantially. To overcome this limitation, we developed \textbf{AloLab}, an iterative system-prompt optimizer (meta-agent: Claude Sonnet 4.5) requiring only black-box API access to the target model; it reaches 84--87\% output accuracy on GSM8K and 34--40\% on MATH across five independent runs per model, with 29/30 paired McNemar comparisons against the best static prompt significant at $p < 0.05$, at near-NAIVE inference latency and without model fine-tuning. The same format failure extends to GPT-4o~\citep{openai2024gpt4o}, a proprietary closed-source model: REFERENCE achieves 0\% output accuracy due to systematic markdown-fence wrapping, while AloLab reaches 95.2\% [94.8, 95.6]. An ablation replacing the Sonnet 4.5 meta-agent with Claude 3 Haiku reduces mean output accuracy to 61.0\% and increases run-to-run standard deviation from $<$1\,pp to 21.8\,pp, confirming that meta-agent capability is a primary driver of optimization quality. A persistent format-execution inconsistency---correct reasoning present but answer field wrong---is independently verified by an LLM judge and accounts for 1.5--1.8\% of remaining AloLab failures.
\end{abstract}

\vspace{0.5em}
\noindent\textbf{Keywords:} Structured output reliability; Automatic prompt optimization; Small language models; JSON format compliance; Constrained decoding; Mathematical reasoning; Black-box LLM optimization.

\section{Introduction}

In production settings, a language model output must satisfy two conditions simultaneously: it must be correct, and it must be parsable by the downstream system. A response that solves the task but violates the output schema is as unusable as one that is simply wrong. Standard benchmarks measure task accuracy alone and do not capture this failure mode. Recent work has characterized this structured-output reliability gap across format-constrained reasoning~\citep{tam2024letmespeakfreely}, format following in 7B-scale models~\citep{wang2025verifiableformat}, JSON schema compliance~\citep{geng2025jsonschemabench}, and structured information extraction~\citep{tenckhoff2026llmstructbench}.

We use mathematics as a stress-test domain for the broader problem of structured-output reliability. The choice is deliberate: mathematical benchmarks provide unambiguous, automatically verifiable ground truth, and the required output format---JSON with fixed fields---imposes a strict, machine-interpretable contract. This makes failures visible and measurable in a way that subjective tasks cannot. The goal is not to advance mathematical reasoning per se, but to study and improve a model's ability to simultaneously solve a task and satisfy an output contract.

We define \textbf{output accuracy} as the joint event of mathematical correctness and valid JSON structure. We evaluate it under five prompting strategies (Section~\ref{sec:configs}) across three 7--9B models on two datasets. A near-universal format failure emerges at baseline: NAIVE prompting yields 0\% output accuracy despite substantial task accuracy. Constrained decoding enforces syntactic validity but introduces large latency overhead and, in several settings, reduces task performance substantially.

We introduce \textbf{AloLab}, an iterative system-prompt optimization framework we developed, requiring only black-box access to the target model---observing input-output behavior via API only, without weight access, gradients, or internal activations. AloLab extends the paradigm of automatic prompt optimization~\citep{yang2023opro, pryzant2023autoprompt, zhou2023ape} to the specific problem of structured-output reliability. AloLab substantially closes the output accuracy gap at near-baseline inference latency. Improvements are statistically robust and consistent across models and datasets.

\section{Method}

\subsection{Models and datasets}

We evaluate three open models: Llama 3.1-8B Instruct~\citep{dubey2024llama3}, Gemma 2-9B IT~\citep{gemmateam2024gemma2}, and Qwen 2.5-7B Instruct~\citep{yang2024qwen2}. Models are served via cloud inference endpoints (Vertex AI / Model Garden vLLM, BF16); no local fine-tuning is performed. These 7--9B models represent a widely deployed tier for cost-sensitive inference~\citep{srivastava2025reasoningslm}, making structured-output reliability especially critical for this family.

To test whether the format failure and AloLab's resolution generalize beyond the small-model regime, we additionally evaluate GPT-4o~\citep{openai2024gpt4o} on GSM8K under three of the five strategies defined in Section~\ref{sec:configs}: NAIVE, REFERENCE, and ALOLAB. GPT-4o is queried via its inference API at temperature~0.0; CONSTRAINED and REF+CONSTRAINED are not evaluated for this model, as they fall outside the scope of this generalization probe.

We use two datasets:
\begin{itemize}
    \item \textbf{GSM8K}~\citep{cobbe2021gsm8k}: natural-language arithmetic word problems.
    \item \textbf{MATH}~\citep{hendrycks2021math}: advanced mathematics across 7 subjects and 5 difficulty levels.
\end{itemize}

For both datasets, we impose a structured JSON output contract as a deliberate research choice: models must respond with a JSON object containing a \texttt{"reasoning"} field and an \texttt{"answer"} field (\texttt{number} for GSM8K; \texttt{\textbackslash boxed\{...\}} for MATH). This contract is not inherent to the datasets---it reflects a realistic deployment requirement and allows us to measure format compliance independently from mathematical correctness.

Splits are fixed and disjoint. GSM8K: optimization $n=150$, validation $n=100$, test $n=1{,}319$. MATH: optimization $n=350$ (stratified by subject$\times$level), validation $n=150$, test $n=5{,}000$. Inference uses temperature 0.0; maximum tokens are 512 (GSM8K) and 2048 (MATH).

\subsection{Metrics}

\textbf{task accuracy}: whether the model's answer is mathematically correct, measured regardless of output format. To allow fair comparison across strategies, we recover the answer from any available signal: from the JSON answer field when the output is valid, from a \texttt{\textbackslash{}boxed\{...\}} expression for MATH, or from the last numerical value in the response for GSM8K. Task accuracy is therefore non-zero even when \texttt{json\_valid} = 0---which is why NAIVE prompting shows high task accuracy despite 0\% output accuracy. For MATH, the evaluator treats mathematically equivalent representations as correct regardless of notation---e.g., \texttt{1/2} and $\frac{1}{2}$ are scored identically, so notation changes introduced during prompt optimization do not affect task accuracy.

\textbf{json\_valid}: output can be parsed as valid JSON with all required fields present.

\textbf{output accuracy}: \texttt{task\_accuracy} $\times$ \texttt{json\_valid}. This is the primary metric throughout. We adopt this term to reflect the deployment perspective: a response that is mathematically correct but structurally invalid cannot be consumed by a downstream system, making it operationally equivalent to a wrong answer. Output accuracy captures this joint condition and distinguishes it from task accuracy, which measures mathematical correctness independently of format compliance.

\subsection{Experimental Configurations}
\label{sec:configs}

We evaluate five experimental configurations:

\textbf{NAIVE}: no system prompt; the model receives only the task-specific user input, with no formatting, behavioral, or output-structure instructions of any kind.

\textbf{REFERENCE}: a minimal hand-written prompt providing basic JSON format instructions, written without iterative refinement or model-specific tuning---the kind of prompt a practitioner would write before running the target model even once, without access to its output behavior. REFERENCE is deliberately kept simple: it represents a realistic static baseline and serves as the starting point from which the AloLab optimization loop begins.

\textbf{CONSTRAINED}: constrained decoding using a JSON grammar (vLLM)~\citep{willard2023outlines, ugare2024syncode}, which enforces syntactically valid JSON at generation time without additional prompt instructions (GSM8K only).

\textbf{REF+CONSTRAINED}: the REFERENCE prompt combined with constrained decoding (GSM8K only). In this configuration, output structure is enforced by the grammar rather than by prompt instructions; the REFERENCE component does not include explicit format directives, which are instead delegated to the constrained decoding layer.

CONSTRAINED and REF+CONSTRAINED are evaluated on GSM8K only; extension to MATH, which requires a more complex answer format inside the JSON field, is left for future work.

\textbf{ALOLAB}: an iterative system-prompt optimizer that uses the target model's own behavior as a feedback signal to progressively rewrite the system prompt (see \S\ref{sec:alolab}). Five independent runs are executed per model and dataset to account for optimization variance.

\begin{figure}[H]
    \centering
    \includegraphics[width=\linewidth]{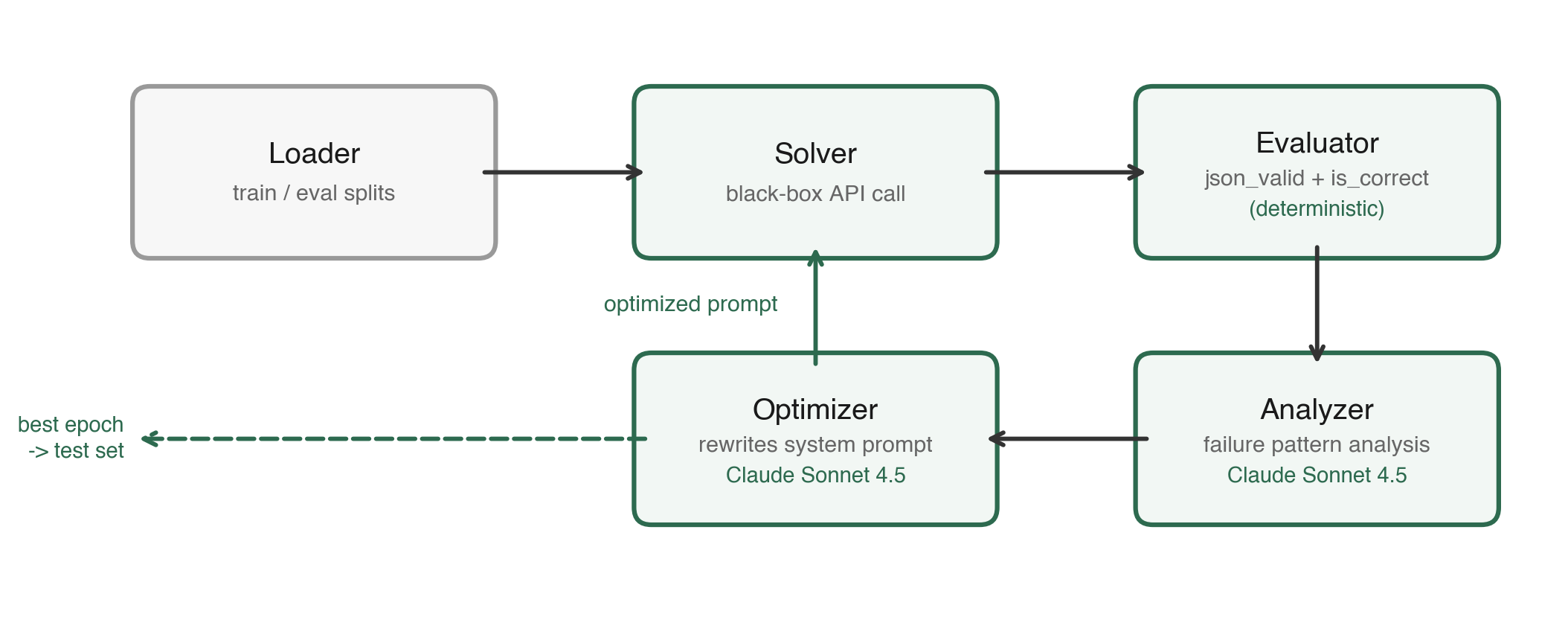}
    \caption{AloLab iterative prompt optimization pipeline. Each epoch: the Solver queries the target model (black-box API only); the Evaluator computes \texttt{json\_valid} and \texttt{is\_correct} deterministically; the Analyzer identifies recurring failure and success patterns; the Optimizer rewrites the system prompt accordingly. After four epochs, the highest-validation checkpoint is selected for test evaluation.}
    \label{fig:architecture}
\end{figure}

\subsection{AloLab}
\label{sec:alolab}

AloLab is an iterative system-prompt optimizer that requires only black-box access to the target model (Figure~\ref{fig:architecture}). Both the Analyzer and Optimizer use Claude Sonnet 4.5~\citep{anthropic2025claude45} as the underlying LLM. The pipeline executes five nodes in sequence each epoch. The Loader prepares the training and evaluation splits. The Solver queries the target model via inference API---no weights, gradients, or internal states are accessed. The Evaluator computes per-sample metrics deterministically: \texttt{json\_valid} and \texttt{is\_correct}. The Analyzer receives all execution traces---both correct and incorrect responses---along with per-sample metrics and a minimal model card (name, architecture family, parameter count, and quantization level). This model card gives the Analyzer the opportunity to account for model-specific behavior: small models can exhibit failure patterns that differ not only across architecture families, but also from what one might expect from a larger model---they may be more sensitive to prompt phrasing, more prone to specific formatting pathologies, or less able to maintain instruction compliance across complex outputs. We wanted to test whether making this information explicit gives the Analyzer and Optimizer a better chance of identifying and addressing these model-specific issues. The Analyzer identifies recurring failure and success patterns, producing structured findings with categorized failure modes and recommendations. The Optimizer receives the current system prompt, the Analyzer's structured findings (including success patterns to preserve), the full optimization history, and the model card; it rewrites the system prompt to address identified failure patterns without disrupting what already works.

Each run starts with an initial prompt (epoch 1) and performs three optimization steps (epochs 2--4). The prompt achieving maximum validation output accuracy is selected for final test evaluation. We execute 5 independent runs per model/dataset and report run means and 95\% bootstrap confidence intervals ($n=10{,}000$ resamples). The iterative optimization loop follows the paradigm introduced by~\citet{yang2023opro} and extended in subsequent work~\citep{pryzant2023autoprompt, zhou2023ape, khattab2024dspy}; a key distinction is that AloLab requires no labeled demonstrations or program-level annotations---only the inference API and a binary correctness signal.

\textbf{Computational cost.} Five independent runs of four epochs each require approximately 5,000--10,000 target-model inference calls (150--350 optimization-split samples + 100--150 validation samples per epoch, for GSM8K and MATH respectively) and 30 Analyzer/Optimizer calls to Claude Sonnet 4.5 per model per dataset. This cost is paid once at optimization time; deployment inference uses only the resulting prompt with no additional overhead.

\textbf{History-only ablation.} To test whether explicit model knowledge is necessary for effective optimization, we run a variant in which neither the Analyzer nor the Optimizer receives any model profile information. Both nodes operate on the same execution traces and aggregate metrics as the full AloLab pipeline, but without knowledge of the target model's identity, architecture family, or quantization level. This isolates whether the iterative feedback loop alone is sufficient, or whether model-specific tailoring is a decisive factor.

\textbf{Meta-agent quality ablation.} To test whether the capability of the meta-agent (Analyzer + Optimizer) is a significant factor, we replace Claude Sonnet 4.5 with Claude 3 Haiku~\citep{anthropic2024claude3haiku} as the underlying LLM for both nodes, keeping all other experimental conditions identical (Llama 3.1-8B Instruct, GSM8K, 5 independent runs). Llama 3.1-8B is selected for this ablation as it showed the lowest run-to-run variance under Sonnet~4.5 (std~$<$1\,pp), providing the cleanest signal for isolating the effect of meta-agent capability. Haiku is a substantially smaller and less capable model than Sonnet 4.5; this ablation isolates whether the iterative optimization loop provides value independent of optimizer quality, or whether meta-agent capability is a primary driver.

\section{Related Work}
\label{sec:relatedwork}

\textbf{Structured-output reliability.} Recent work has characterized the gap between correctness and format compliance across several settings: format-constrained reasoning~\citep{tam2024letmespeakfreely}, format following in 7B-scale models~\citep{wang2025verifiableformat}, JSON schema compliance~\citep{geng2025jsonschemabench}, and structured information extraction~\citep{tenckhoff2026llmstructbench}. Our work studies this gap in a controlled mathematical setting where ground truth is unambiguous and the output contract is strict, enabling independent measurement of task accuracy and format validity.

\textbf{Constrained decoding.} Grammar-based constrained decoding~\citep{willard2023outlines, ugare2024syncode} enforces syntactic validity at generation time. Reducing its overhead is an active research area~\citep{park2024grammaraligned, sun2025earleydriven}. Hybrid frameworks~\citep{nguyen2026thinkingbefore} allow models to reason freely before switching to constrained generation, directly targeting the reasoning-format tension. Constrained decoding has also been extended to diffusion language models~\citep{mundler2025constraineddiffusion}. We show that even optimized constrained decoding incurs substantial per-inference overhead and, in some models, degrades task performance.

\textbf{Automatic prompt optimization.} AloLab extends the paradigm of iterative prompt optimization~\citep{yang2023opro, pryzant2023autoprompt, zhou2023ape} to the specific problem of structured-output reliability. Related frameworks include DSPy~\citep{khattab2024dspy} and memory-augmented approaches that accumulate cross-task optimization strategies~\citep{liang2026memapo}. A key distinction is that AloLab requires no labeled demonstrations, program-level annotations, or gradient access---only black-box API calls and a binary correctness signal.

\textbf{Reinforcement learning for structured output.} Reinforcement learning approaches train models to reason over structured API calls without explicit format supervision~\citep{jin2025searchr1, ye2026incontextrl}. These methods require weight access and training infrastructure; AloLab achieves structured-output reliability through prompt optimization alone, with no model modifications.

\section{Results}

NAIVE, REFERENCE, CONSTRAINED, and REF+CONSTRAINED are each evaluated in a single run (deterministic decoding at temperature~0.0). AloLab is evaluated across five independent runs per model and dataset; we report means and 95\% bootstrap confidence intervals.

\subsection{Format gap}

Table~\ref{tab:naive} reports NAIVE performance. All models achieve non-trivial task accuracy but 0\% output accuracy on both datasets: without formatting guidance, no model produces valid JSON. The task accuracy range spans 77--85\% on GSM8K and 4--69\% on MATH for the three open-weight models, confirming that the bottleneck is format, not reasoning. GPT-4o reaches 84.3\% task accuracy under NAIVE on GSM8K, also with 0\% output accuracy.

\begin{table}[H]
\centering
\caption{NAIVE baseline: task accuracy vs.\ output accuracy across models and datasets.}
\label{tab:naive}
\begin{tabular}{llccc}
\toprule
Dataset & Model & Task Acc. & json\_valid & Out. Acc. \\
\midrule
GSM8K & Llama 3.1-8B Instruct & 76.88\% & 0.0\% & 0.0\% \\
      & Gemma 2-9B IT         & 80.44\% & 0.0\% & 0.0\% \\
      & Qwen 2.5-7B Instruct  & 85.1\%  & 0.0\% & 0.0\% \\
      & GPT-4o$^{\star}$      & 84.3\%  & 0.0\% & 0.0\% \\
\midrule
MATH  & Llama 3.1-8B Instruct & 4.06\%  & 0.0\% & 0.0\% \\
      & Gemma 2-9B IT         & 10.36\% & 0.0\% & 0.0\% \\
      & Qwen 2.5-7B Instruct  & 69.22\% & 0.0\% & 0.0\% \\
\bottomrule
\end{tabular}
\smallskip\\
{\footnotesize $^{\star}$GPT-4o evaluated on GSM8K only (generalization probe); MATH not evaluated for this model.}
\end{table}

\subsection{All-strategy comparison}

Tables~\ref{tab:gsm8k} and~\ref{tab:math} show output accuracy across all strategies. On GSM8K, AloLab dominates in every case. REFERENCE reaches 74.3\% for Qwen but fails entirely for Gemma, which systematically wraps JSON in markdown fences---rendering all outputs unparseable despite 88.4\% underlying task accuracy (see Appendix~\ref{app:prompts} for the prompt evolution). CONSTRAINED enforces syntactic validity but at high latency cost and with quality degradation for Gemma (52.4\% of outputs are exact duplicates under this strategy). On MATH, REFERENCE is near zero for two of three models; AloLab is the only strategy that reliably crosses 30\%.

\begin{table}[H]
\centering
\caption{GSM8K output accuracy (\%) under all strategies. AloLab reports mean over 5 runs with 95\% bootstrap CI.}
\label{tab:gsm8k}
\begin{tabular}{lccccc}
\toprule
Model & NAIVE & REF & CONST & REF+CONST & ALOLAB \\
\midrule
Llama 3.1-8B Instruct & 0.0 & 44.43 & 52.46 & 19.56 & \textbf{84.15} \small{[83.71, 84.59]} \\
Gemma 2-9B IT   & 0.0 & 0.0$^{\dagger}$ & 15.31 & 22.90 & \textbf{87.41} \small{[87.10, 87.70]} \\
Qwen 2.5-7B Instruct  & 0.0 & 74.30 & 32.83 & 46.93 & \textbf{85.75} \small{[80.00, 88.79]} \\
\midrule
GPT-4o$^{\star}$  & 0.0 & 0.0$^{\dagger\dagger}$ & --- & --- & \textbf{95.22} \small{[94.75, 95.59]} \\
\bottomrule
\end{tabular}
\smallskip\\
{\footnotesize $^{\dagger}$Gemma REFERENCE json\_valid = 0\%: the model wraps every JSON response in markdown fences (\texttt{```json...```}); task accuracy recovered by the fallback parser is 88.4\%, but all outputs are unparseable.\\
$^{\dagger\dagger}$GPT-4o REFERENCE json\_valid = 0\%: the same markdown-fence wrapping observed in Gemma---a model-specific generation default that the static REFERENCE prompt does not suppress; task accuracy recovered by the fallback parser is 95.7\%.\\
$^{\star}$GPT-4o is a proprietary closed-source model evaluated as a generalization probe; CONSTRAINED strategies were not evaluated for this model.}
\end{table}

\begin{table}[H]
\centering
\caption{MATH output accuracy (\%) under all strategies. AloLab reports mean over 5 runs with 95\% bootstrap CI. CONSTRAINED and REF+CONSTRAINED were not evaluated on MATH.}
\label{tab:math}
\begin{tabular}{lccc}
\toprule
Model & NAIVE & REFERENCE & ALOLAB \\
\midrule
Llama 3.1-8B Instruct & 0.0 & 8.54  & \textbf{40.02} \small{[38.35, 42.02]} \\
Gemma 2-9B IT   & 0.0 & 0.0$^{\ddagger}$ & \textbf{40.24} \small{[36.12, 42.68]} \\
Qwen 2.5-7B Instruct  & 0.0 & 0.0$^{\ddagger}$ & \textbf{34.27} \small{[20.83, 47.71]}$^{\S}$ \\
\bottomrule
\end{tabular}
\smallskip\\
{\footnotesize $^{\ddagger}$Gemma and Qwen REFERENCE MATH: json\_valid = 0\% (markdown fence and format non-compliance, respectively); task accuracy recovered by the fallback parser is 32.0\% (Gemma) and 47.5\% (Qwen).\\
$^{\S}$Wide CI reflects genuine run-to-run variance: two Qwen runs produce double-escaped LaTeX (\texttt{\textbackslash\textbackslash boxed\{\}}) that the parser does not recognize, yielding near-zero output accuracy despite high task accuracy in those runs.}
\end{table}

\subsection{Constrained decoding trade-offs}

\begin{figure}[H]
    \centering
    \includegraphics[width=0.85\linewidth]{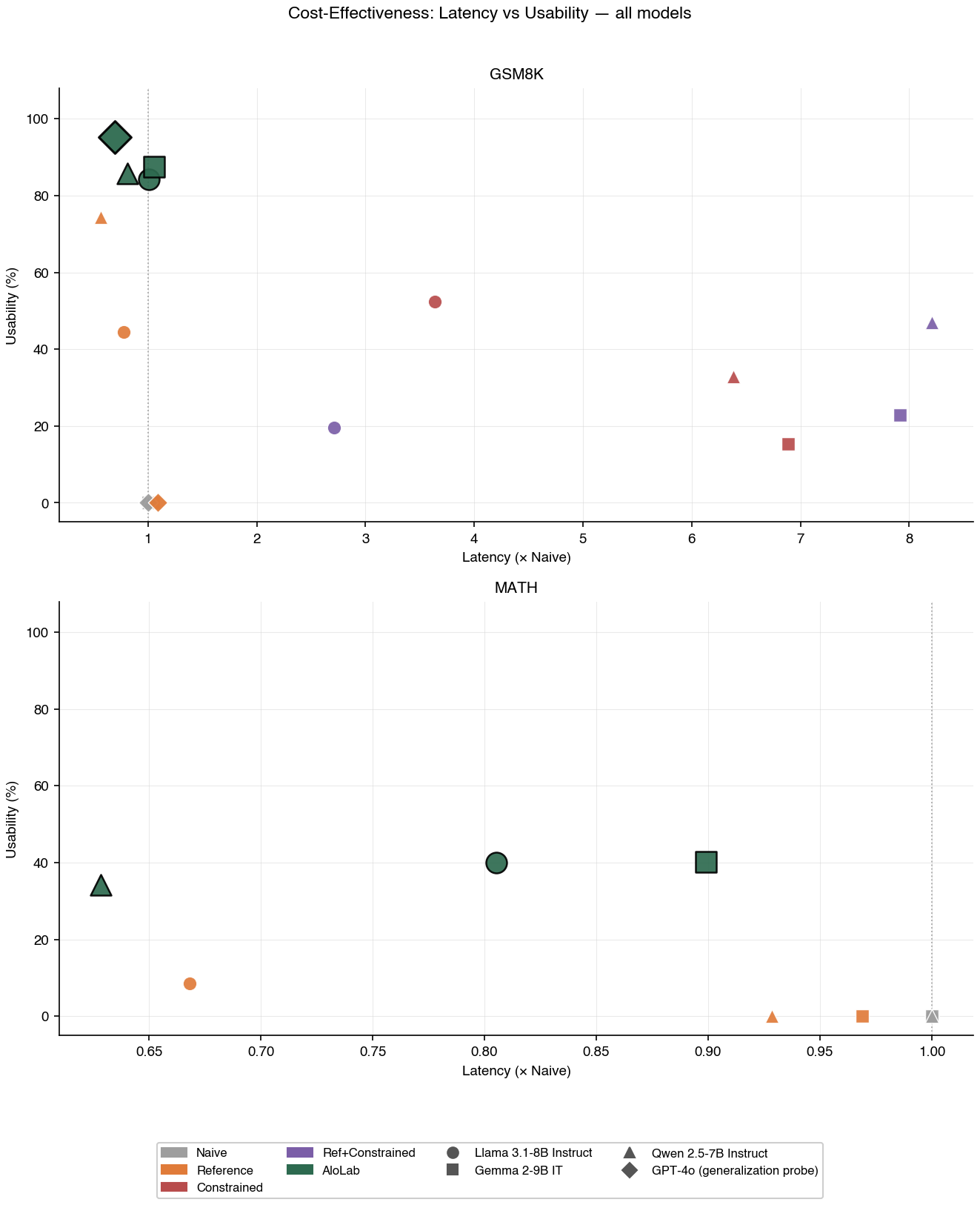}
    \caption{Cost-effectiveness: output accuracy vs.\ relative inference latency (normalized to NAIVE) for GSM8K (top) and MATH (bottom). AloLab achieves top output accuracy at near-NAIVE latency; CONSTRAINED occupies the high-latency, lower-output-accuracy region. Diamond markers ($\Diamond$) denote GPT-4o (generalization probe, GSM8K only): NAIVE and REFERENCE at 0\% output accuracy, AloLab at 95.2\% and 0.69$\times$ NAIVE latency. Note the different x-axis scales: GSM8K spans up to 8$\times$ Naive (CONSTRAINED overhead), while MATH strategies all remain below 1$\times$.}
    \label{fig:pareto}
\end{figure}

Constrained decoding enforces syntactic validity at substantial cost. On GSM8K, inference latency relative to NAIVE is 3.6$\times$ (Llama CONSTRAINED) to 8.2$\times$ (Qwen REF+CONSTRAINED). Despite this overhead, output accuracy remains below AloLab for all models. Reducing constrained-decoding overhead is an active research area~\citep{park2024grammaraligned, sun2025earleydriven}, but even optimized implementations incur per-token cost at every inference call, whereas prompt-level optimization pays the cost once. AloLab inference latency is 0.71$\times$--1.06$\times$ that of NAIVE on GSM8K and 0.63$\times$--0.90$\times$ on MATH---shorter optimized prompts partially offset token overhead. Figure~\ref{fig:pareto} shows the cost-effectiveness frontier. In concrete terms: a deployment serving responses at AloLab latency would require 3.6--8.2$\times$ longer wait under CONSTRAINED---at lower output accuracy.

\subsection{Convergence dynamics}
\label{sec:convergence}

\begin{figure}[H]
    \centering
    \includegraphics[width=\linewidth]{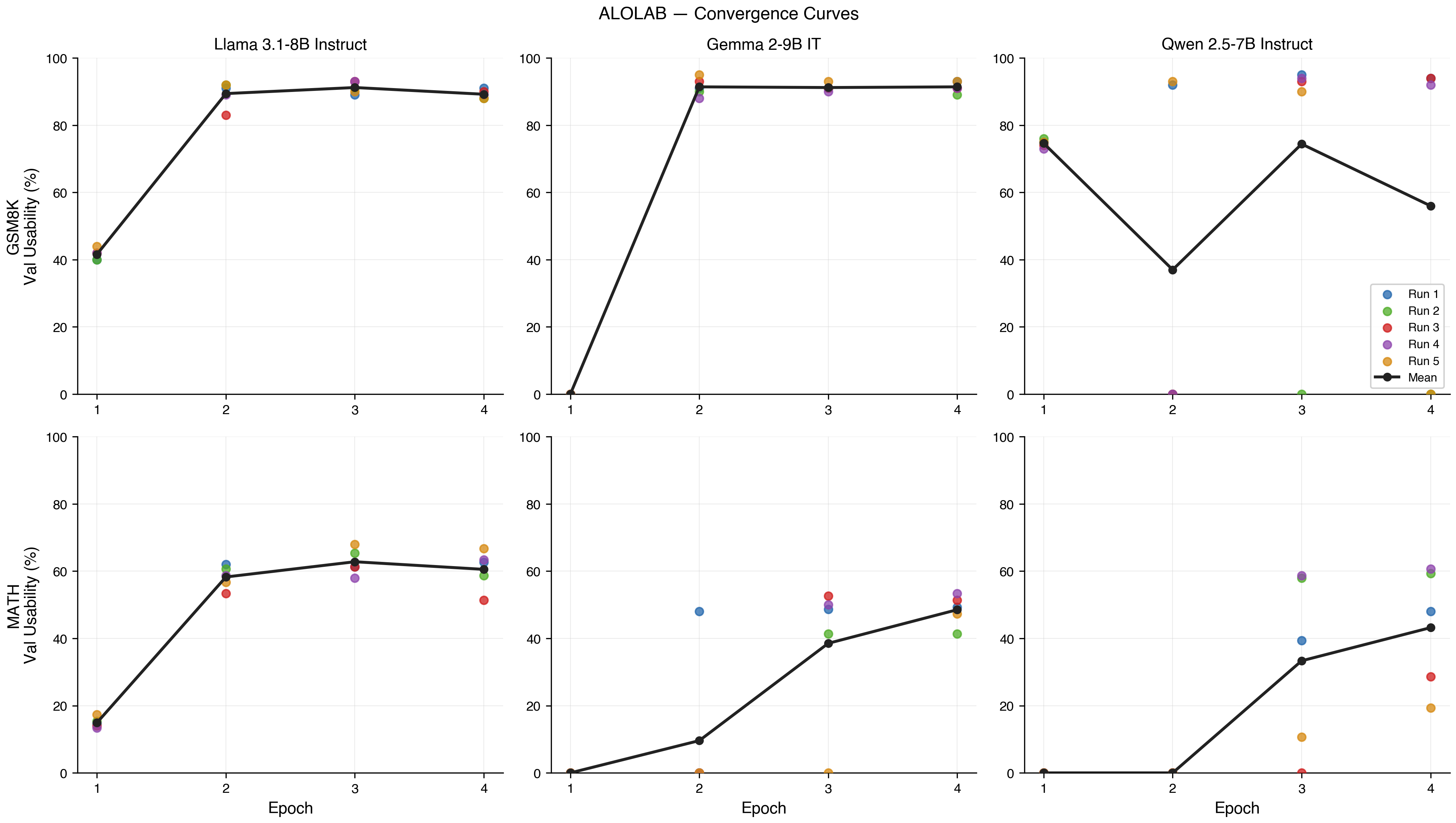}
    \caption{Validation output accuracy over optimization epochs for all models and datasets (GSM8K top row, MATH bottom row). Each line is one independent run.}
    \label{fig:convergence}
\end{figure}

Figure~\ref{fig:convergence} shows validation output accuracy trajectories over optimization epochs. On GSM8K, Llama and Gemma converge reliably within 2--3 epochs. Llama jumps from its REFERENCE baseline to near-ceiling output accuracy at epoch~2. Gemma on GSM8K shows the sharpest transition: epoch-1 output accuracy is 0\% across all five runs (the REFERENCE prompt produces no valid JSON due to markdown-fence wrapping), followed by a single-epoch breakthrough at epoch~2 as the optimizer identifies and suppresses the fence. Qwen on GSM8K converges in most runs but exhibits occasional sharp drops---one run collapses at epoch~2 and never recovers (Section~\ref{sec:significance})---resulting in a less stable mean trajectory and wider confidence intervals than Llama and Gemma. On MATH, convergence is more gradual for all models; Gemma again starts at 0\% output accuracy in epoch~1 (same markdown-fence issue) and climbs progressively across epochs 2--4. Qwen shows occasional degradation trajectories on both datasets---the optimizer takes a suboptimal direction in some runs---which accounts for the wider confidence intervals for Qwen in Tables~\ref{tab:gsm8k} and~\ref{tab:math}.

\subsection{Difficulty and subject stratification}

\begin{figure}[H]
    \centering
    \includegraphics[width=\linewidth]{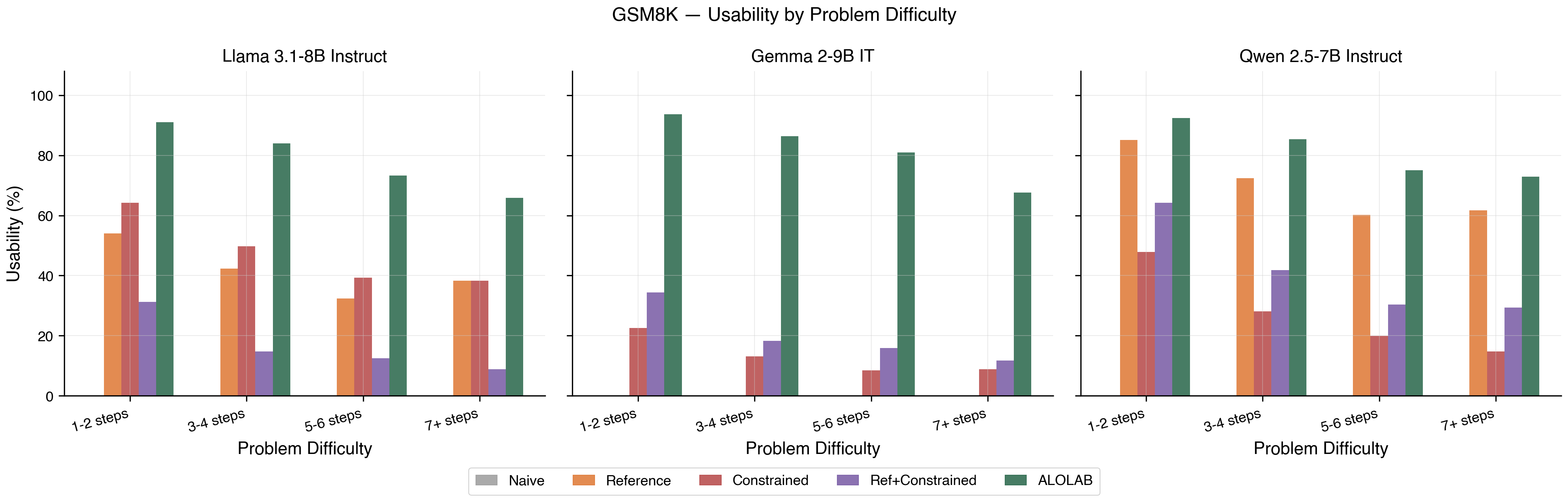}
    \caption{GSM8K output accuracy stratified by estimated solution complexity (number of reasoning steps), shown per model under all strategies.}
    \label{fig:gsm8k_strat}
\end{figure}

\begin{figure}[H]
    \centering
    \includegraphics[width=\linewidth]{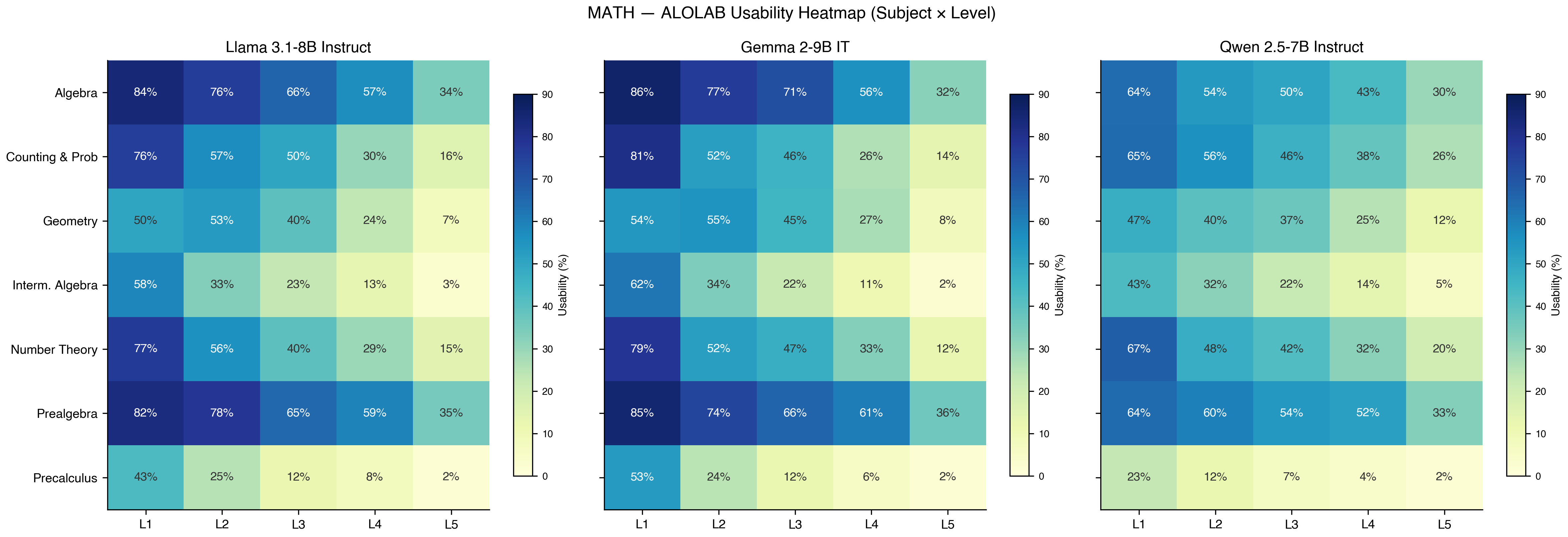}
    \caption{AloLab MATH output accuracy heatmap by subject and difficulty level for each model. Performance degrades with level (columns) and shifts across subjects (rows) consistently across all three models.}
    \label{fig:math_heatmap}
\end{figure}

Figures~\ref{fig:gsm8k_strat} and~\ref{fig:math_heatmap} show performance stratified by problem difficulty. For GSM8K, we estimate solution complexity by counting the explicit calculation annotations (\texttt{<<...>>}) in each ground-truth solution and bucketing into four groups (1--2, 3--4, 5--6, 7+ steps). Output accuracy declines monotonically with this estimate: from 91.2\% at 1--2 reasoning steps to 66.7\% at 7+ steps (averaged across models). On MATH, the decline is steeper: from 67.1\% at Level 1 to 17.9\% at Level 5. Subject-wise, Prealgebra (60.4\%) and Algebra (58.7\%) are strongest; Intermediate Algebra (25.2\%) and Precalculus (15.7\%) are weakest. These gradients are consistent across all three models, indicating they reflect intrinsic task difficulty rather than any model-specific artifact.

\subsection{Statistical significance}
\label{sec:significance}

We report paired McNemar tests (AloLab vs.\ REFERENCE on matched samples). Of 30 run-level comparisons (5 runs $\times$ 3 models $\times$ 2 datasets), 29 are significant at $p < 0.05$ (all with $p \approx 0$). The single exception is Qwen GSM8K run 2 ($p = 0.286$): this run undergoes a JSON collapse at epoch 2, json\_valid drops to near zero and never recovers, and the resulting test output accuracy (74.3\%) is statistically indistinguishable from REFERENCE (74.3\%). This is a documented convergence failure (Section~\ref{sec:convergence}), not evidence of no effect---the other 4/5 Qwen GSM8K runs are all significant.

\subsection{Generalization to GPT-4o}
\label{sec:gpt4o}

The format failure is not confined to small open-weight models. GPT-4o, a proprietary closed-source model, produces 0\% output accuracy under both NAIVE and REFERENCE prompting on GSM8K: without a system prompt the model outputs free-form prose incompatible with our JSON extractor; with the REFERENCE prompt it systematically wraps JSON in markdown fences---the same failure mode observed in Gemma 2-9B. AloLab resolves it identically: by observing the wrapping behavior in the optimization traces, the Optimizer produces an explicit no-fence directive, raising output accuracy to \textbf{95.22\%} [94.75, 95.59] across five independent runs (std $= 0.55$\,pp). This result confirms that the format failure is a model-specific generation default---not a size-related limitation---and that black-box iterative optimization addresses it regardless of model scale.

\subsection{Meta-agent quality ablation}
\label{sec:haiku_ablation}

\begin{figure}[H]
    \centering
    \includegraphics[width=\linewidth]{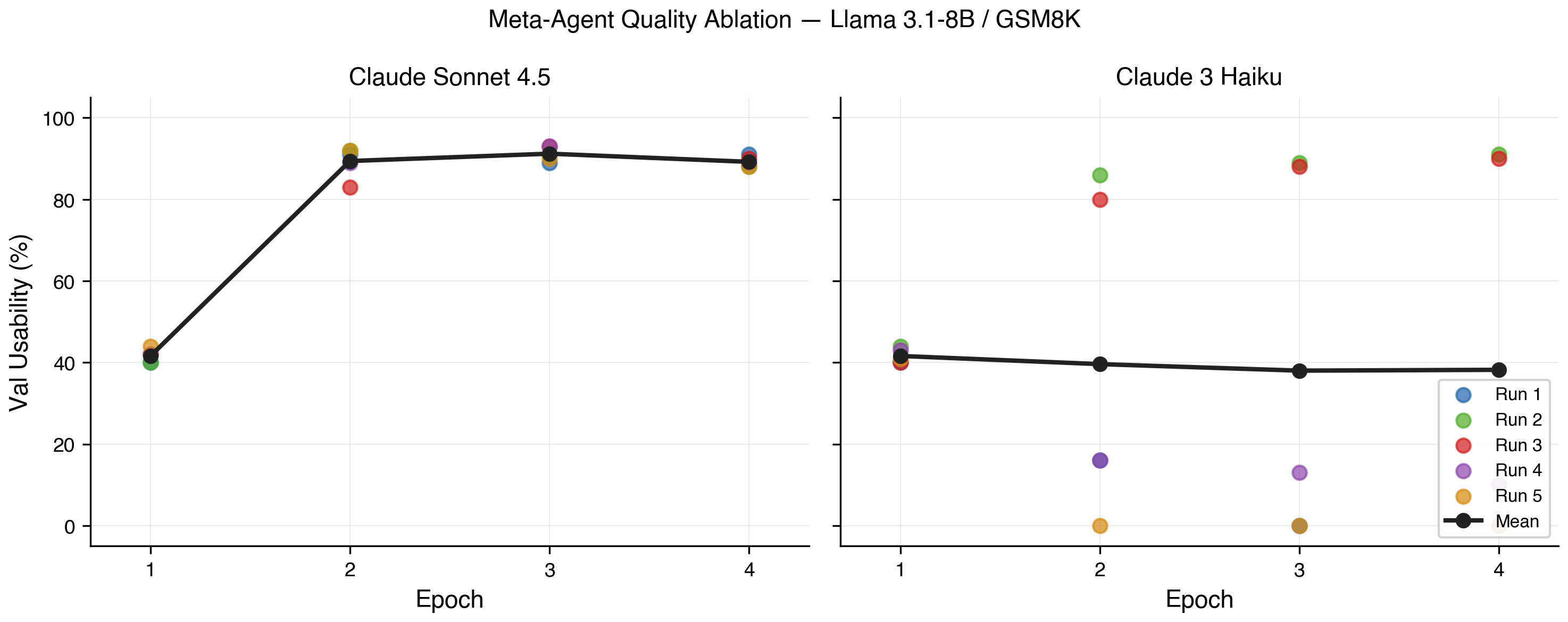}
    \caption{Meta-agent quality ablation: validation output accuracy per epoch for Llama 3.1-8B / GSM8K under Claude Sonnet 4.5 (left) vs.\ Claude 3 Haiku (right) as meta-agent. Each line is one independent run. Sonnet 4.5 converges reliably across all runs; Haiku produces a bimodal distribution---two runs find an effective prompt, three fail or collapse.}
    \label{fig:haiku_ablation}
\end{figure}

Figure~\ref{fig:haiku_ablation} and Table~\ref{tab:haiku_ablation} show the effect of replacing the Sonnet 4.5 meta-agent with Claude 3 Haiku on Llama 3.1-8B / GSM8K. The weaker optimizer achieves a mean output accuracy of 61.0\%---23\,pp below Sonnet 4.5 (84.15\%)---and, critically, exhibits dramatically higher run-to-run variance (std $= 21.8$\,pp vs.\ $< 1$\,pp for Sonnet). Three of five Haiku runs converge to low-quality prompts ($< 50$\% output accuracy), while two discover effective prompts comparable to Sonnet's best. This bimodal distribution indicates that a weak meta-agent can occasionally find a good solution but cannot do so reliably: the optimization loop provides value in principle, but meta-agent capability determines whether that value is realized consistently.

\begin{table}[H]
\centering
\caption{Meta-agent quality ablation on Llama 3.1-8B / GSM8K (5 runs each). Output accuracy (\%) per run and aggregate statistics.}
\label{tab:haiku_ablation}
\begin{tabular}{lccccccc}
\toprule
Meta-agent & Run 1 & Run 2 & Run 3 & Run 4 & Run 5 & Mean & Std \\
\midrule
Claude Sonnet 4.5 & 83.78 & 84.76 & 84.61 & 84.23 & 83.40 & 84.15 & 0.6\,pp \\
Claude 3 Haiku    & 45.49 & 84.38 & 85.44 & 44.43 & 45.26 & 61.00 & 21.8\,pp \\
\bottomrule
\end{tabular}
\end{table}

\section{Discussion}

\textbf{Latency and cost-effectiveness.}
AloLab is both more effective and faster at inference time than CONSTRAINED. CONSTRAINED imposes a 3.6$\times$--8.2$\times$ overhead at \emph{every} inference call, which is expensive to scale. AloLab incurs its optimization cost once; subsequent inference uses only the optimized prompt, with no decoding overhead. Notably, on MATH the optimized prompt achieves inference latency \emph{below} NAIVE (0.63$\times$--0.90$\times$): the optimizer tends to produce more directive system prompts that elicit more concise completions---an effect absent in constrained decoding, which adds overhead at every inference call. The Pareto view (Figure~\ref{fig:pareto}) makes this trade-off explicit.

\textbf{What drives the optimization.}
Removing the model profile from both the Analyzer and Optimizer---while keeping all other inputs (execution traces, per-sample metrics, and optimization history) unchanged---shifts mean output accuracy by $-0.7$ pp on average (Llama: $-0.3$ pp, Gemma: $-0.6$ pp, Qwen: $-1.3$ pp), with no statistically significant effect. Explicit model knowledge does not provide a measurable advantage: the optimization signal is captured by observed execution patterns, regardless of whether the optimizer is informed of the target model's identity, architecture, or quantization level.

A related question is whether the optimizer's pre-training knowledge of these benchmarks explains the gains. Appendix~\ref{app:prompts} provides concrete evidence against this interpretation: the key interventions---suppressing markdown fences for Gemma on GSM8K, and switching from LaTeX to plain-text notation for Llama on MATH---are responses to observed model behavior, not predictions derivable from prior knowledge of the benchmarks. The REFERENCE baseline represents a simple static starting point---the kind of prompt a practitioner would write without observing model behavior. The gap between AloLab and REFERENCE ranges from $-11$ pp (Qwen) to $-87$ pp (Gemma). For Gemma, no amount of static benchmark knowledge could produce the required fix: the optimizer must \emph{observe} that this specific model systematically wraps JSON in markdown fences, and write a prompt that prevents it. That behavior is not predictable from prior knowledge of the task. This distinguishes the contribution of the iterative loop from any confound of optimizer pre-training.

\textbf{Format-task tension.}
Enforcing a strict output format can impose a modest cost on task accuracy. On Qwen MATH, the task accuracy recovered by the fallback parser under REFERENCE---where json\_valid is 0\%---is 47.5\%, while AloLab mean output accuracy is 34.27\%. This gap reflects a format-task tension: directing the model to produce structured output constrains its generation in ways that can slightly reduce underlying task performance. The net effect is nonetheless strongly positive---output accuracy improves from 0\% to 34.27\%---and is consistent with the broader observation that output token format affects answer quality in a model-dependent way~\citep{hamilton2025lostinspace}. The wide confidence interval for Qwen MATH ([20.8\%, 47.7\%]) compounds this effect, as runs that converge to incompatible formats contribute near-zero output accuracy despite high task accuracy.

\textbf{Format-execution inconsistency.}
Across all AloLab runs, a fraction of failures belong to a specific category: json\_valid = True (the output is a parseable JSON object) but is\_correct = False (the \texttt{answer} field holds the wrong value)---yet reading the \texttt{reasoning} field reveals that the model was in fact working toward the correct answer. The model reasons correctly but writes the wrong value into the \texttt{answer} field: the \texttt{reasoning} and \texttt{answer} fields decouple. A regex scan of the \texttt{reasoning} field places a rough upper bound of 15--21\% on this category, but over-counts due to incidental matches (e.g., the target number appearing as an intermediate calculation result). An independent LLM judge (Gemini 3 Flash;~\citealt{google2025gemini3flash}) provided a tighter estimate: given the \texttt{reasoning} field and the ground truth, the judge assessed whether the model was semantically concluding the correct answer, independent of any formatting. The verified true rate is 1.5--1.8\% of AloLab failures in this pool---far lower than the regex bound suggests. For context, the same judge estimated rates of 0--0.97\% for CONSTRAINED and 0--0.16\% for REFERENCE---configurations where valid-JSON output is either rarer or syntactically enforced. The higher rate in AloLab is consistent with this being an intrinsic generation property that becomes more visible as prompt optimization eliminates simpler format errors: the residual failures are disproportionately those where reasoning and answer-field serialization decouple. \citet{hamilton2025lostinspace} show that the choice of output token format has systematic, model-dependent effects on answer quality, suggesting this decoupling is a broader phenomenon beyond our setting. Addressing it is a concrete direction for future work.

\section{Limitations}

\textbf{Optimizer dependence.} AloLab uses Claude Sonnet 4.5~\citep{anthropic2025claude45} as its meta-agent. The Haiku ablation (Section~\ref{sec:haiku_ablation}) shows that a weaker optimizer substantially degrades both mean performance and consistency: mean output accuracy drops from 84.15\% to 61.0\% and standard deviation increases from $<$1\,pp to 21.8\,pp. The optimization loop retains value in principle---two of five Haiku runs find effective prompts---but reliability requires a capable meta-agent. Whether mid-tier models achieve intermediate trade-offs remains open. AloLab also incurs a one-time optimization cost (see \S\ref{sec:alolab} for details) that is a meaningful setup investment for scenarios requiring rapid deployment or frequent task switching.

\textbf{Benchmark scope.} Both datasets are mathematical. This is a deliberate choice: mathematical benchmarks provide controlled, verifiable conditions where correctness and format compliance are independently measurable. Generalization to other task types and output schemas remains to be established and is a direct next step for future work. While contamination concerns have been raised for both GSM8K and MATH~\citep{mirzadeh2024gsmsymbolic}, our evaluation focuses on format compliance and our comparisons are relative and within-distribution, making contamination unlikely to confound the main conclusions.

\textbf{Context window.} For MATH, we use a fixed context window of 8{,}192 tokens across all models, which is shorter than the native context of some models and may impose an artificial ceiling.

\textbf{Ablation scope.} The history-only ablation and the meta-agent quality ablation were both conducted on GSM8K only; MATH results may differ given the more complex formatting requirements (LaTeX in JSON). The GPT-4o generalization probe covers GSM8K only; MATH and multi-model frontier results remain to be established.

\textbf{Qwen MATH variance.} The wide Qwen MATH confidence interval [20.8\%, 47.7\%] reflects genuine instability in the optimization landscape: even a capable optimizer can converge to syntactically plausible but parser-incompatible patterns (double-escaped LaTeX in two of five runs). The double-escaped pattern (\texttt{\textbackslash\textbackslash boxed\{\}}) is reproducible without AloLab: Qwen generates it under any prompt that permits LaTeX in the answer field, including REFERENCE. AloLab eliminates it in 3/5 runs but cannot do so consistently, suggesting the instability is intrinsic to Qwen's generation behavior rather than a failure of the optimization loop. Reporting five independent runs with bootstrap confidence intervals is therefore essential for capturing this genuine instability rather than masking it with a single point estimate.

\section{Conclusion}

We study structured-output reliability in small language models under strict JSON output contracts, using mathematical benchmarks as a controlled testbed where correctness and format compliance are independently measurable. A systematic format failure emerges at baseline: high task accuracy does not imply output accuracy when format requirements are strict. Constrained decoding addresses syntax but introduces latency overhead and can degrade task performance. AloLab, an iterative black-box prompt optimizer, substantially closes the output accuracy gap without fine-tuning, with statistically robust gains at near-baseline inference cost. A persistent format-execution inconsistency remains an open challenge. These results suggest that the practical limits of small language models~\citep{srivastava2025reasoningslm} under strict output contracts are not limits of reasoning---NAIVE task accuracy reaches 77--85\% on GSM8K---but limits of structured communication. Iterative black-box prompt optimization directly targets this gap and is both more accurate and less costly at inference time than constrained decoding.

\bibliographystyle{abbrvnat}
\bibliography{references}

\appendix

\section{Prompt Evolution Examples}
\label{app:prompts}

We show two representative optimization trajectories that illustrate qualitatively different failure modes and the strategies AloLab develops to address them.

\subsection{Gemma 2-9B IT on GSM8K: eliminating markdown wrapping}

\paragraph{Initial prompt (epoch~1 / REFERENCE).}
AloLab starts from the same prompt as the REFERENCE baseline---a prompt a human expert might reasonably write:

\begin{quote}
\textit{You are a math expert. Solve the problem step by step, then give the final numeric answer.\\
Respond with a JSON object with keys ``reasoning'' (string) and ``answer'' (number).}
\end{quote}

Under this prompt, every Gemma output follows the pattern:

\begin{verbatim}
```json
{
  "reasoning": "Janet uses 3 eggs for breakfast and 4 for muffins,
                leaving 9 to sell at $2 each: 9 x $2 = $18.",
  "answer": 18
}
```
\end{verbatim}

The answer is mathematically correct, but the markdown fence makes the response unparseable as JSON. json\_valid = 0\% across all 1,319 test samples.

\paragraph{AloLab epoch~2.}
After observing epoch-1 failures, the Optimizer rewrites the prompt. The critical addition:

\begin{quote}
\textit{CRITICAL OUTPUT FORMAT RULES:\\
1. Output ONLY a raw JSON object -- no markdown, no code blocks, no extra text.\\
2. Do NOT wrap your response in \texttt{```json```} or any other formatting.\\
3. The JSON must have exactly two keys: ``reasoning'' (string) and ``answer'' (number).}
\end{quote}

This single intervention raises output accuracy from 0\% to $\approx$80\% on the validation set. No static prompt author could predict this model-specific behavior without observing it; constrained decoding enforces syntax but degrades task accuracy for Gemma under grammar constraints.

\subsection{Llama 3.1-8B Instruct on MATH: changing the answer representation}

MATH introduces a qualitatively different challenge: answers must be mathematical expressions (fractions, radicals, symbolic forms), and the standard notation is LaTeX. When a model writes LaTeX inside a JSON string, unescaped backslashes produce structurally invalid JSON.

\paragraph{Initial prompt (epoch~1 / REFERENCE).}
AloLab starts from the same minimal, instruction-only prompt as the REFERENCE baseline---no few-shot examples:

\begin{quote}
\textit{You are a math expert. Solve the problem step by step, then give the final answer.\\
Respond with a JSON object with keys ``reasoning'' (string) and ``answer'' (string).\\
Your final answer must be a mathematical expression. Use LaTeX notation if needed (e.g.\ \texttt{\textbackslash frac\{1\}\{2\}}, \texttt{\textbackslash sqrt\{3\}}).}
\end{quote}

The third line is a notation hint, not a few-shot example: it instructs the model to write mathematical expressions in LaTeX. Following this instruction literally, a typical output looks like:

\begin{verbatim}
{
  "reasoning": "120% of 30 is 1.2 \times 30 = 36.
                130% of 20 is 1.3 \times 20 = 26.
                Positive difference: 36 - 26 = 10.",
  "answer": "$36 - 26 = 10$"
}
\end{verbatim}

The sequence \verb|\t| (from \verb|\times|) is an invalid JSON escape; the parser rejects the object. With the REFERENCE prompt, json\_valid = 47.4\% and output accuracy = 8.5\%.

\paragraph{AloLab epoch~2.}
Rather than instructing the model to escape backslashes---a correction easy to specify but hard to apply reliably---the Optimizer takes the opposite approach: it eliminates LaTeX from the answer field entirely.

\begin{quote}
\textit{ANSWER FORMAT RULES:\\
-- For fractions: use ``a/b'' format (e.g.\ ``3/4'')\\
-- For square roots: use ``sqrt(n)'' format (e.g.\ ``sqrt(5)'')\\
-- For expressions: use plain text algebra (e.g.\ ``x\^{}2 + 2x + 1'')}
\end{quote}

With no backslashes to escape, json\_valid rises to 77.9\% and output accuracy to 39.0\%---a $4.6\times$ improvement over REFERENCE that no amount of static prompt refinement could reach, because the LaTeX requirement itself was the source of the problem.

\medskip
\noindent Together, these examples illustrate that AloLab does not apply a fixed correction: it identifies the specific failure mode of the target model and finds the most direct path to resolve it---whether that is a wrapping convention or a representation strategy.

\end{document}